\documentclass[sigconf]{acmart}
\usepackage{booktabs}
\usepackage{graphicx}
\usepackage{tabularx}
\usepackage{array}
\providecommand{\Description}[1]{}
\setcopyright{none}
\acmYear{2027}
\copyrightyear{2027}
\acmDOI{}
\acmISBN{}
\newcolumntype{Y}{>{\centering\arraybackslash}X}

\begin{document}
\title{TESR: Target-Aware Early-Stage Ranking through Model--Serving Co-Design}
\author{Juhee Hong}
\authornote{All three authors contributed equally to this research.}
\email{jessiejuheehong@meta.com}
\orcid{1234-5678-9012}
\affiliation{%
  \institution{Meta Platforms, Inc.}
  \city{Menlo Park}
  \state{CA}
  \country{USA}
}

\author{Meng Liu}
\authornotemark[1]
\email{mengliu2019@meta.com}
\affiliation{%
  \institution{Meta Platforms, Inc.}
  \city{Menlo Park}
  \state{CA}
  \country{USA}
}

\author{Shengzhi Wang}
\authornotemark[1]
\email{shengzhi@meta.com}
\affiliation{%
  \institution{Meta Platforms, Inc.}
  \city{Menlo Park}
  \state{CA}
  \country{USA}
  }

\author{Jin Zhou}
\affiliation{%
  \institution{Meta Platforms, Inc.}
  \city{Menlo Park}
  \state{CA}
  \country{USA}
  }

\author{Xiaoheng Mao}
\affiliation{%
  \institution{Meta Platforms, Inc.}
  \city{Menlo Park}
  \state{CA}
  \country{USA}
  }

\author{Zhao Zhu}
\affiliation{%
  \institution{Meta Platforms, Inc.}
  \city{Menlo Park}
  \state{CA}
  \country{USA}
  }

\author{Ruochen Liu}
\affiliation{%
  \institution{Meta Platforms, Inc.}
  \city{Menlo Park}
  \state{CA}
  \country{USA}
  }

\author{Huihui Cheng}
\affiliation{%
  \institution{Meta Platforms, Inc.}
  \city{Menlo Park}
  \state{CA}
  \country{USA}
  }

\author{Leon Gao}
\affiliation{%
  \institution{Meta Platforms, Inc.}
  \city{Menlo Park}
  \state{CA}
  \country{USA}
  }

\author{Christopher Leung}
\affiliation{%
  \institution{Meta Platforms, Inc.}
  \city{Menlo Park}
  \state{CA}
  \country{USA}
  }

\author{Chandra Mouli Sekar}
\affiliation{%
  \institution{Meta Platforms, Inc.}
  \city{Menlo Park}
  \state{CA}
  \country{USA}
  }

\author{Yijia Liu}
\affiliation{%
  \institution{Meta Platforms, Inc.}
  \city{Menlo Park}
  \state{CA}
  \country{USA}
  }

\author{Boyang Yu}
\affiliation{%
  \institution{Meta Platforms, Inc.}
  \city{Menlo Park}
  \state{CA}
  \country{USA}
  }

\author{Rui Li}
\affiliation{%
  \institution{Meta Platforms, Inc.}
  \city{Menlo Park}
  \state{CA}
  \country{USA}
  }

\author{Jing Qian}
\affiliation{%
  \institution{Meta Platforms, Inc.}
  \city{Menlo Park}
  \state{CA}
  \country{USA}
  }

\author{Xuan Cao}
\affiliation{%
  \institution{Meta Platforms, Inc.}
  \city{Menlo Park}
  \state{CA}
  \country{USA}
  }

\author{Minjie Fan}
\affiliation{%
  \institution{Meta Platforms, Inc.}
  \city{Menlo Park}
  \state{CA}
  \country{USA}
  }

\author{Mingze Gao}
\affiliation{%
  \institution{Meta Platforms, Inc.}
  \city{Menlo Park}
  \state{CA}
  \country{USA}
  }

\renewcommand{\shortauthors}{Hong, Liu, Wang et al.}
\begin{abstract}
Early-stage ranking (ESR) systems must score hundreds to thousands
of candidates under tight serving constraints. Production systems
therefore commonly decouple user and item computation through
two-tower architectures, improving throughput but limiting
fine-grained interaction between each candidate and the user's
interaction history. We present \textbf{Target-Aware Early-Stage
Ranking (TESR)}, a model--serving co-design that makes
candidate-dependent sequence modeling practical in ESR. TESR
integrates two complementary interaction paths: Exact ID Matching
(EIM) across engagement-specific histories and
candidate-conditioned cross-attention over encoded interaction
sequences. Its serving design encodes history once per request,
confines additional repeated work to candidate-dependent operators,
and executes those operators using fused kernels and ahead-of-time
graph compilation. TESR Basic reduces relative normalized entropy
by 0.75\% and 0.40\% on engagement and consumption tasks,
respectively; controlled ablations show that removing either
interaction path worsens both tasks. In production-derived replay,
Basic sustains 3.1\% more requests than the production two-tower
baseline under the evaluated serving constraints. Two controlled
A/B experiments show positive online lifts for TESR Basic over
Production Two Tower and for a launch package centered on TESR
Advanced over Basic across engagement, consumption, and overall
product activity. Together, these results provide evidence that
candidate-conditioned sequence interaction can operate at ESR scale
through model--serving co-design.
\end{abstract}
\maketitle
\section{Introduction}

\begin{figure}[t]
  \centering
  \includegraphics[width=\columnwidth]{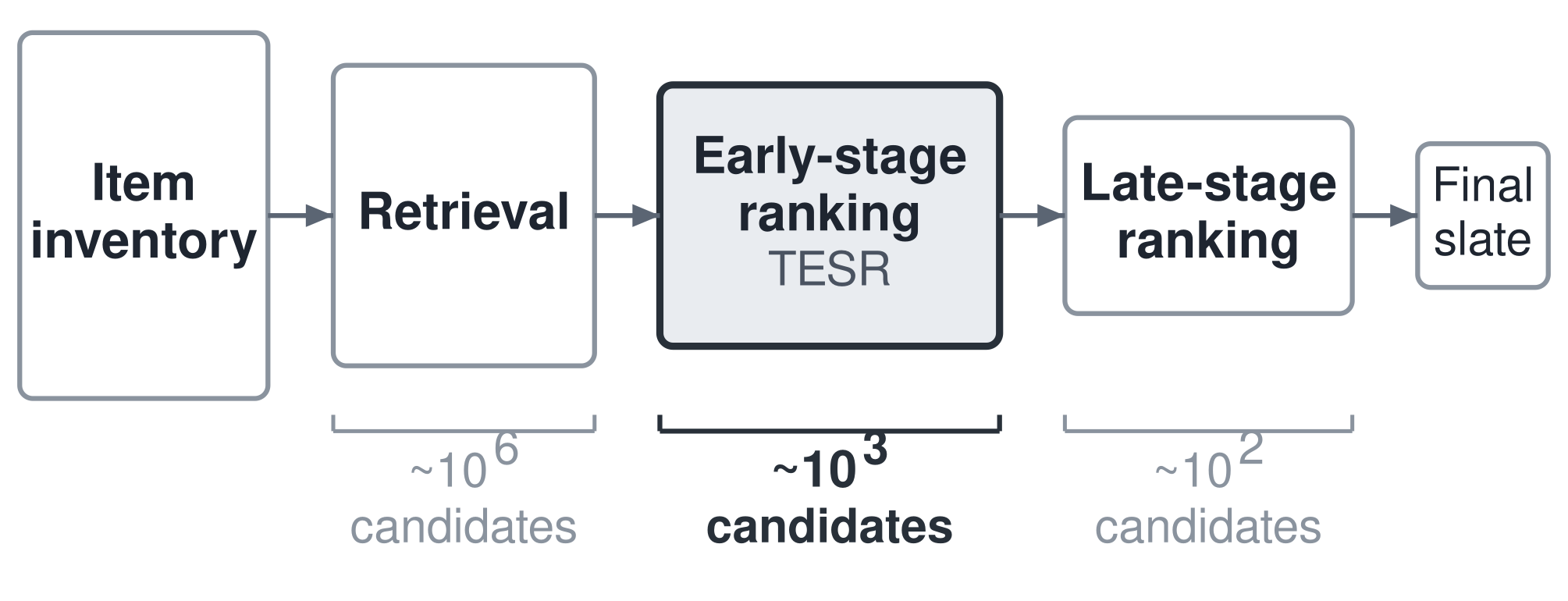}
  \caption{Ranking cascade and candidate counts. ESR's larger pool amplifies
  repeated per-candidate work.}
  \Description{A ranking cascade from item inventory through retrieval,
  early-stage ranking, late-stage ranking, and a final slate.}
  \label{fig:cascade}
\end{figure}

Large-scale recommendation, advertising, and search systems commonly use
multi-stage ranking cascades to balance effectiveness and
efficiency~\cite{covington2016youtube,huang2020embedding}. Retrieval searches a
massive corpus using lightweight methods to produce an initial candidate pool.
Early-stage ranking (ESR) then scores and filters hundreds to thousands of
candidates, passing a much smaller subset to late-stage ranking (LSR) for final
ordering (Figure~\ref{fig:cascade}). Because LSR evaluates fewer candidates, it
can devote more computation to each one; the distinction is one of feasible
compute allocation, not an intrinsic property of the stage. Consequently,
candidate-dependent operations that are affordable in LSR can become
prohibitive when repeated across the larger ESR pool.

\par\indent Two-tower models separate request-side user computation from
precomputable or cached item-side computation. This efficient ESR design
weakens fine-grained candidate--history
interaction~\cite{huang2013dssm,li2022inttower}. We call our deployed reference
Production Two Tower. TESR's novelty is jointly realizing exact matching and
candidate-conditioned Hierarchical Sequential Transduction Unit (HSTU)
attention~\cite{zhai2024hstu} under ESR-scale constraints while retaining
reusable computation.

\par\indent We study the following systems question: \textbf{how can ESR
condition sequence interactions on each candidate while retaining a
production-feasible operating point?} Our answer is \textbf{Target-Aware
Early-Stage Ranking (TESR)}, a co-designed model and serving architecture. TESR
combines exact matching with candidate-conditioned sequence attention, then
bounds and amortizes these operations through request-shared computation,
cacheable candidate features, deferred broadcasting, model pruning,
mixed-precision serving, specialized kernels, and compiler lowering.

\par\indent Our contributions are:
\begin{itemize}
  \item \textbf{Target-aware interaction modeling.} TESR augments Production
  Two Tower with EIM and candidate-conditioned HSTU attention while preserving
  reusable request and item computation.
  \item \textbf{Model--serving co-design.} Request-shared computation,
  cacheable features, deferred broadcasting, pruning, mixed precision, fused
  kernels, and compiler lowering make these interactions practical.
\item \textbf{Quality, serving, and online evidence.} Controlled
offline ablations measure the contribution of both interaction paths;
production-derived replay shows that Basic sustains higher
full-service throughput under the evaluated constraints; and two
controlled A/B experiments show positive online lifts for TESR Basic
and for a launch package centered on TESR Advanced.
\end{itemize}

\begin{figure*}[t]
  \centering
  \includegraphics[width=0.86\textwidth]{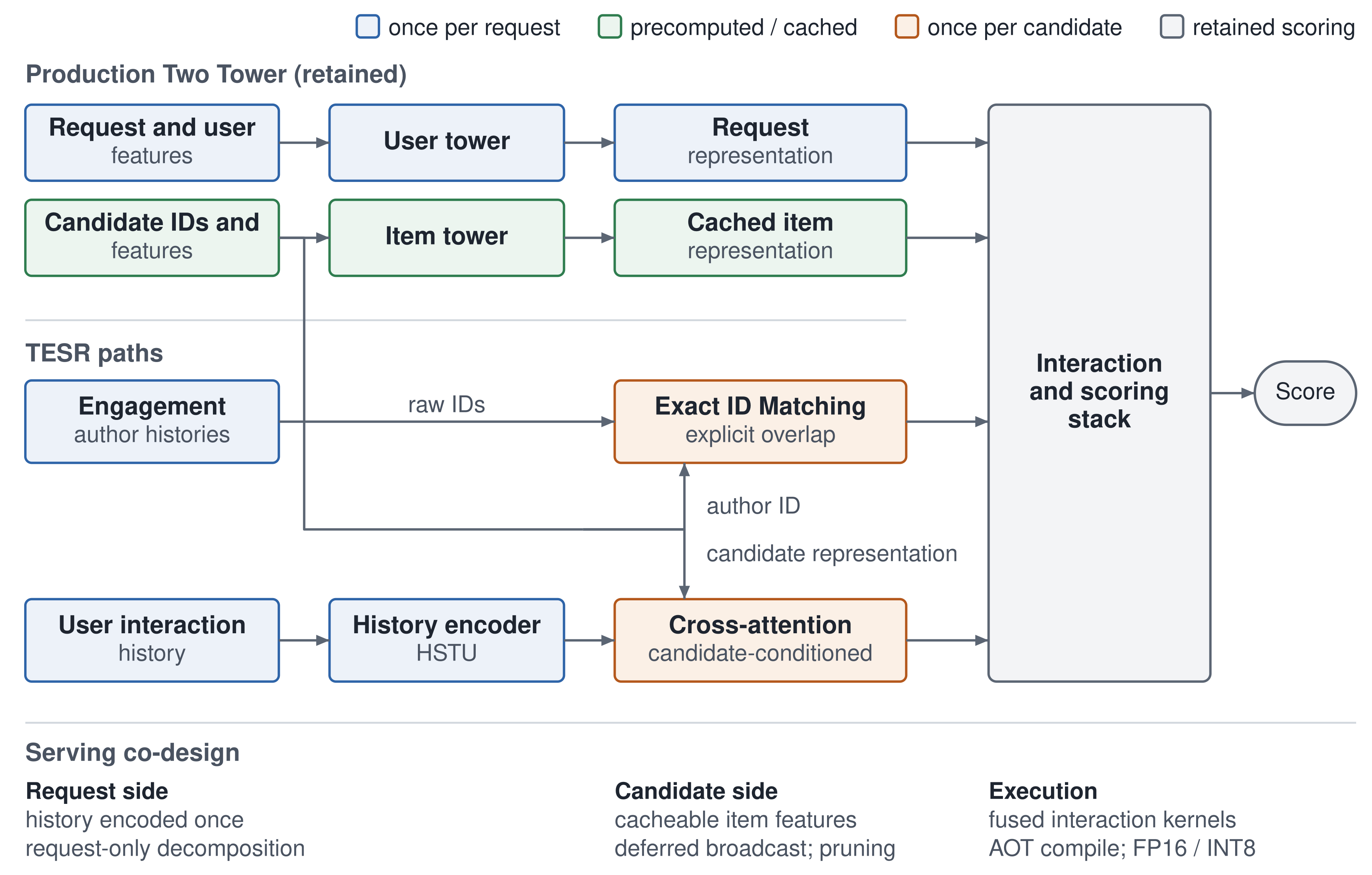}
  \caption{TESR retains Production Two Tower while adding exact matching and
  candidate-conditioned sequence attention.}
  \Description{The Production Two Tower backbone augmented with Exact ID
  Matching and candidate-conditioned sequence attention, grouped by execution
  frequency.}
  \label{fig:architecture}
\end{figure*}
\section{Model--System Co-design}
For each request, TESR scores $N$ candidates against shared history
and context. Production Two Tower computes a request representation
once and combines it with cached candidate representations. TESR
preserves this amortized path while adding request-shared history
encoding and candidate-conditioned interaction paths
(Figure~\ref{fig:architecture}).

\subsection{Complementary Interaction Paths}
\par
\begin{figure*}[t]
  \centering
  \includegraphics[width=0.86\textwidth]{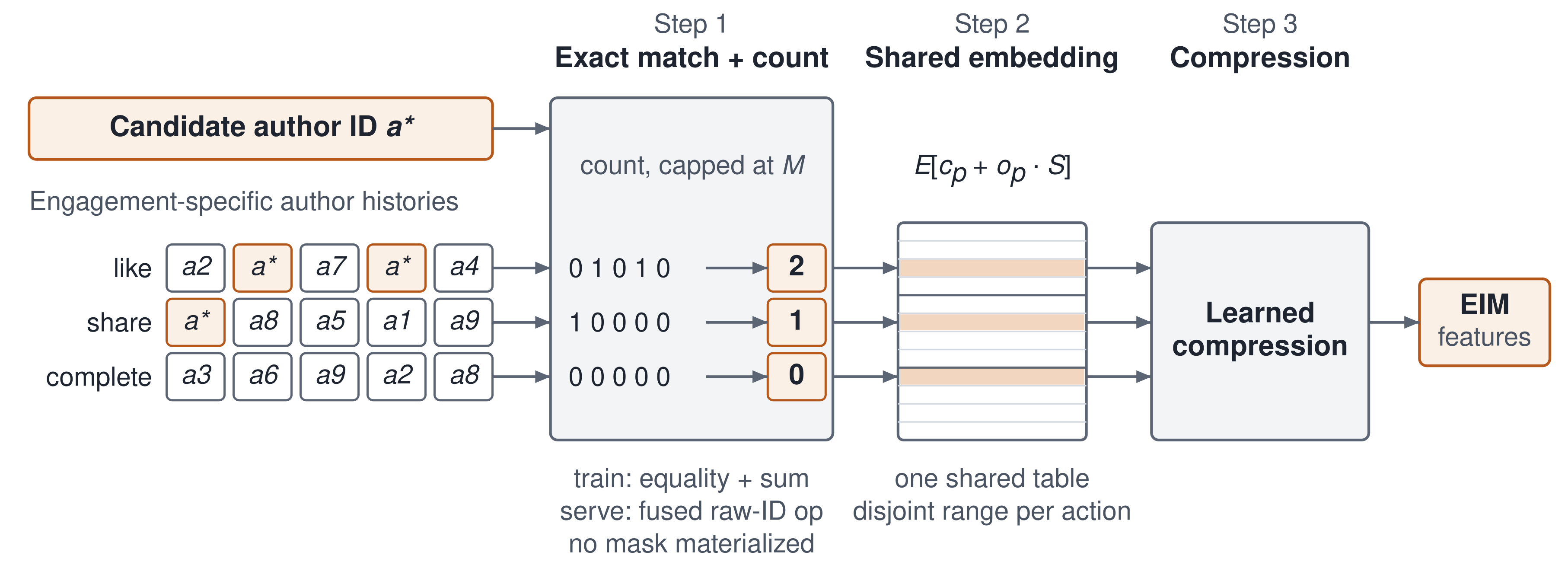}
  \caption{Exact ID Matching instantiated with author IDs. Capped
per-action candidate--history match counts index disjoint ranges of
a shared embedding table, and learned compression combines the
resulting embeddings.}
  \Description{Three-step candidate-author ID matching, offset-indexed shared count embedding, and learned compression pipeline.}
  \label{fig:eim}
\end{figure*}

\indent\textbf{Exact ID Matching.}
EIM captures exact categorical overlap signals that can be obscured
by dense two-tower representations. The operator can use any
categorical key available for both candidates and historical events;
our evaluated TESR configuration uses author ID. For each engagement
action---such as like, comment, save, share, follow, profile visit,
or completed or non-skipped view---the corresponding history stores
the author IDs of previously engaged items. EIM compares each
candidate's author ID with every action-specific history and caps
the resulting match count $c_p$ for action $p$ at $M$
(Figure~\ref{fig:eim}, Step~1). Conceptually, each count reduces a
sequence of binary exact-match indicators, although serving does
not materialize this mask. A shared table then embeds
$E[c_p+o_pS]$, where $o_p$ selects an action-specific range with
stride $S=M+1$ (Step~2), and a learned compressor combines the
per-action embeddings (Step~3). Training computes equivalent counts
through tensor equality and reduction, whereas serving uses a fused
raw-ID operator vectorized over the candidate batch. This design
avoids materialized crossed features and makes exact candidate-history
interaction practical at ESR scale.
\par
\begin{figure}[t]
  \centering
  \includegraphics[width=\columnwidth]{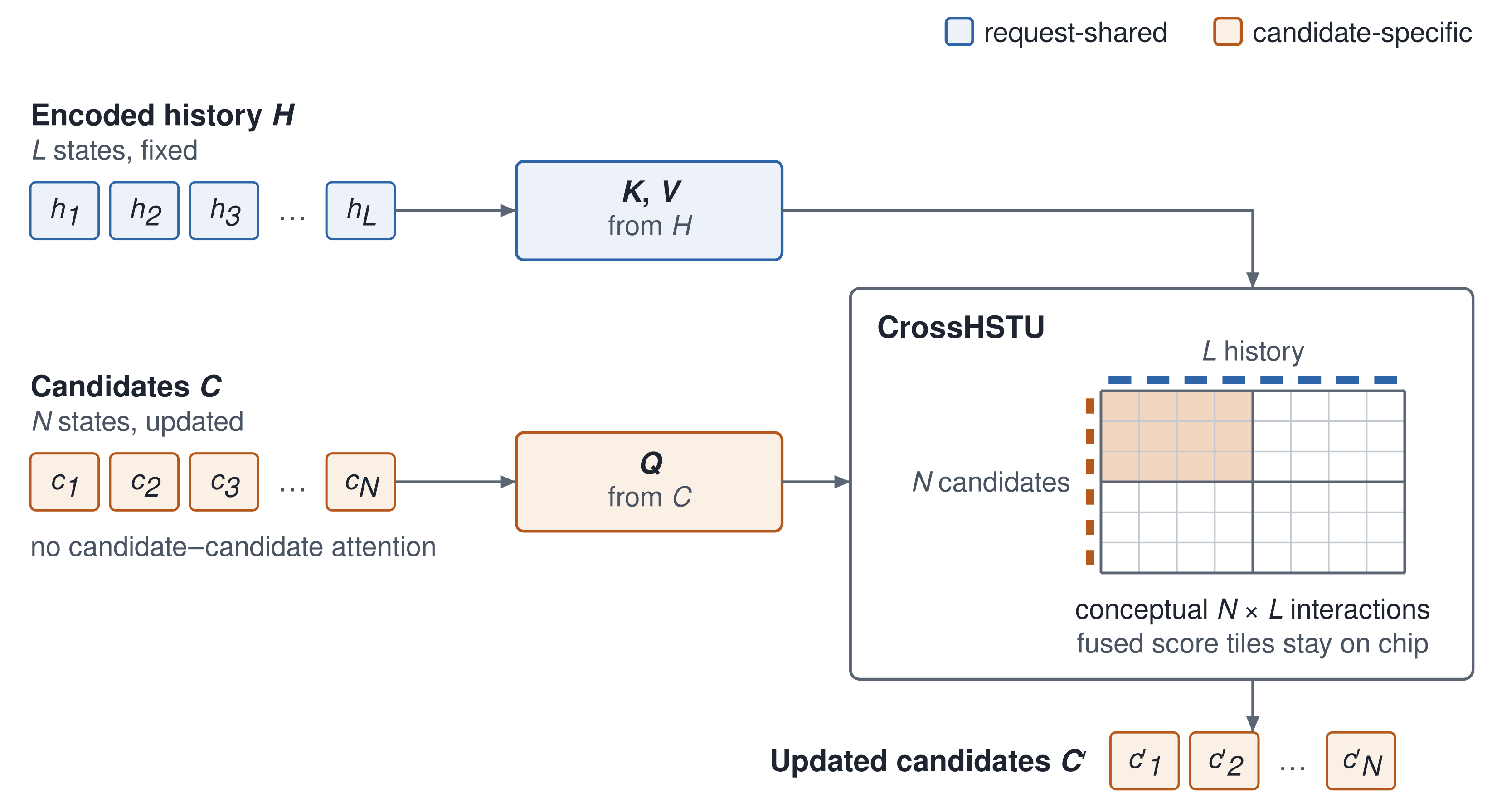}
  \caption{Candidate-conditioned cross-attention uses candidate queries and
history keys/values to update only candidate states before scoring.}
  \Description{Candidate queries and fixed history key-value projections enter
an HSTU-derived cross-attention operator whose fused score tiles update only
candidate representations.}
  \label{fig:cross-attention}
\end{figure}

\indent\textbf{Cross-attention.}
Cross-attention provides learned target awareness: each candidate queries encoded user interaction history to retrieve evidence relevant to that target, including latent affinities that do not require exact identifier overlap~\cite{pi2020sim,chang2023twin,si2024twinv2}. Let $H\in\mathbb{R}^{L\times d}$ denote encoded history and $C\in\mathbb{R}^{N\times d}$ candidate representations derived from media and author identifiers. HSTU-derived layers~\cite{zhai2024hstu} project candidate states into queries $Q=CW_Q$ and history into keys and values $K=HW_K$, $V=HW_V$. Each layer produces an $N\times L$ target--history interaction map and updates only candidate states; history remains fixed and candidates do not attend to one another (Figure~\ref{fig:cross-attention}). This concentrates computation on target--history evidence without recomputing history-side outputs.
\subsection{Model--Serving Co-design}
\par\indent\textbf{Execution-frequency factorization.}
Serving cost depends on execution multiplicity as well as operator
complexity. Following the decomposition in
Figure~\ref{fig:architecture}, TESR serving cost is
\begin{equation}
  C_{\mathrm{TESR}}
  = C_{\mathrm{2T}}(N)
  + C_{\mathrm{hist}}(L)
  + C_{\mathrm{EIM}}(N,\{L_p\})
  + C_{\mathrm{attn}}(N,L).
\end{equation}
Here $C_{\mathrm{2T}}(N)$ denotes the retained Production Two Tower
path, including request-side computation, cached item
representations, and per-candidate scoring; its dimensions depend
on the Basic or Advanced configuration. The TESR history encoder
$C_{\mathrm{hist}}(L)$ runs once per request. For action-history
lengths $\{L_p\}$, EIM costs $O(N\sum_p L_p)$, while
candidate-conditioned attention costs $O(NLd)$. Thus TESR encodes
history once, retains cacheable item computation, and confines its
new candidate--history interactions to EIM and cross-attention.
\par\indent\textbf{Fused interaction kernels.}
At serving time, EIM fuses raw-ID comparison and capped counting in
an operator vectorized over the candidate batch. 

For cross-attention, $Q$ and $K,V$ are stored as unequal packed
streams with per-request offsets. Unlike softmax attention, the
HSTU-derived operator applies pointwise score activation and
per-sequence scaling. A custom Triton kernel~\cite{tillet2019triton}
tiles the $N\times L$ interaction space, activates each score tile,
and immediately accumulates it against $V$. Only one tile remains
on chip before reduction to the $O(Nd)$ output, and the $O(NL)$
interaction matrix is never written to global memory. Arithmetic
remains $O(NLd)$; the kernel reduces padding and intermediate
traffic rather than making candidate--history interaction
sublinear.
\par\indent\textbf{Capacity reallocation.}
The higher-fidelity interaction signal lets TESR reduce redundant capacity elsewhere. Basic downsizes embedding tables and trims the surrounding Production Two Tower stack, including redundant user--item projections; Advanced additionally halves the user-tower hidden dimension. This shifts capacity toward candidate--history interaction and reduces whole-model floating-point operations (FLOPs) despite adding attention.
\par\indent\textbf{Ahead-of-time graph compilation.}
High-dimensional request representations remain request-shared until the candidate merge, while compatible embedding and interaction dimensions reduce redundant projections. PyTorch Inductor~\cite{ansel2024pytorch2} compiles the remaining dense graph ahead of deployment, fusing pointwise operations and matrix-multiplication epilogues and removing framework dispatch. Candidate count and history length remain symbolic dimensions, allowing one compiled artifact to support variable request shapes.
\par\indent\textbf{Mixed-precision lowering.}
Precision is selected by operator type rather than applied uniformly. Training uses BF16 in eligible dense and sequence modules. Serving uses FP16 dense execution and INT8 embedding tables; candidate-interaction operators use dedicated inference kernels while Inductor lowers the surrounding dense graph.
\section{Evaluation}
\label{sec:evaluation}
\subsection{Offline Evaluation}
\par\indent\textbf{Data and scope.}
Our evaluation uses proprietary, production-scale data from a deployed
short-video ESR system. Because TESR couples model and serving changes, to our
knowledge no public benchmark exposes both the required histories and ESR
execution environment; we therefore combine controlled offline ablations,
production-derived replay, and randomized online experiments. All offline
configurations use the same examples, features, labels, and protocol.
Production Two Tower, which computes separate request and item representations
and introduces user--item interaction only in the final scoring stack, is the
reference. Basic uses both interactions with a four-head, 256-dimensional HSTU
encoder over 1,024 events; Advanced uses 512 dimensions and 2,048 events.
Normalized entropy (NE) is log loss divided by empirical label entropy; lower
is better. Table~\ref{tab:offline} reports relative percentage changes, so more
negative values indicate better predictions. E-task represents engagement
(e.g., reshare) and C-task consumption (e.g., completed viewing). The
\emph{Basic +} rows independently isolate longer history and wider encoding;
Advanced includes both scale-ups.
\begin{table}[t]
  \caption{Relative NE (\%) vs. Production Two Tower. The ``+'' rows isolate one scale-up; lower is better.}
  \label{tab:offline}
  \centering
  \footnotesize
  \setlength{\tabcolsep}{3pt}
  \begin{tabularx}{\columnwidth}{@{}Xcc@{}}
    \toprule
    Setup & E-task NE & C-task NE \\
    \midrule
    Basic w/o cand.-conditioned seq. & $-0.20\%$ & $-0.24\%$ \\
    Basic w/o EIM & $-0.68\%$ & $-0.33\%$ \\
    \textbf{Basic} & \textbf{$-0.75\%$} & \textbf{$-0.40\%$} \\
    Basic + 2k history & $-0.87\%$ & $-0.47\%$ \\
    Basic + 2$\times$ encoder & $-0.94\%$ & $-0.54\%$ \\
    \textbf{Advanced} & \textbf{$-1.10\%$} & \textbf{$-0.60\%$} \\
    \bottomrule
  \end{tabularx}
\end{table}
\par\indent\textbf{Results.}
Ablating either candidate-conditioned sequence modeling or EIM consistently degrades both engagement and consumption performance relative to TESR Basic. Neither mechanism subsumes the other: EIM contributes high-precision identity overlap, while candidate-conditioned sequence modeling captures latent affinity beyond exact matches. Independently increasing sequence length or encoder width improves both tasks further, and the combined Advanced endpoint performs best.
\subsection{Model and Serving Efficiency}
We evaluate efficiency at two scopes.
Table~\ref{tab:model-efficiency} reports training queries per second
(QPS), total-model FLOPs, and model-only inference QPS relative to
Production Two Tower. Table~\ref{tab:service-efficiency} evaluates
Basic, the efficiency-oriented configuration, for maximum sustainable
service throughput under limits of 80\,ms 99th-percentile (p99)
processing latency and a 0.1\% failure rate. The baseline and Basic
runs use comparable production-derived replay workloads averaging
$\sim$800 candidates per request, together with the same serving
harness and constraints. QPS is comparable only within each protocol;
candidate throughput is measured per GPU card.
\begin{table}[t]
  \caption{Model-level efficiency relative to Production Two Tower; negative
deltas denote reductions.}
  \label{tab:model-efficiency}
  \centering
  \footnotesize
  \setlength{\tabcolsep}{4pt}
  \begin{tabularx}{\columnwidth}{@{}Xcc@{}}
    \toprule
    Metric & Basic & Advanced \\
    \midrule
    Training QPS $\Delta$ & $-17\%$ & $-25\%$ \\
    Total-model FLOPs $\Delta$ & $-12\%$ & $-15\%$ \\
    Model inference QPS $\Delta$ & $-36\%$ & $-43\%$ \\
    \bottomrule
  \end{tabularx}
\end{table}
\begin{table}[t]
  \caption{Basic vs. Production Two Tower on comparable replay
workloads under 80\,ms p99 latency and a 0.1\% failure-rate limit.}
  \label{tab:service-efficiency}
  \centering
  \footnotesize
  \setlength{\tabcolsep}{4pt}
  \begin{tabularx}{\columnwidth}{@{}Xc@{}}
    \toprule
    Metric & Basic \\
    \midrule
    Request QPS $\Delta$ & \textbf{$+3.1\%$} \\
    Candidate throughput $\Delta$ & $+1.8\%$ \\
    \bottomrule
  \end{tabularx}
\end{table}
\par\indent\textbf{Efficiency results.}
Basic and Advanced reduce total-model FLOPs by 12\% and 15\%, respectively.
A negative FLOPs delta denotes a whole-model reduction, not cheaper attention.
Lower arithmetic cost does not directly imply higher throughput: irregular,
memory-intensive sequence operators reduce training and model-only inference
QPS. End to end, Basic nevertheless sustains 3.1\% more requests and
1.8\% more candidates per second per card across the comparable
replay workloads under the shared serving constraints. These service
gains coexist with lower model-only QPS because the full path includes
request-shared encoding, cached item features, embedding-service work, fused
interactions, and compiled graph execution. The result is consistent with the
Section~2.2 co-design: repeated work is confined to fused candidate-dependent
interactions while the remaining dense graph is compiled.
Tables~\ref{tab:model-efficiency} and~\ref{tab:service-efficiency} use
different protocols, so their QPS deltas are not compared directly.
\subsection{Online Results and Deployment}
Table~\ref{tab:online} reports two controlled A/B experiments. The
first compares TESR Basic with Production Two Tower. The second
evaluates a launch package centered on TESR Advanced but also
containing other production changes, using Basic as its control.
We therefore attribute the second result to the launch package
rather than to TESR Advanced alone.

Both experiments produce consistently positive online lifts across
engagement, consumption, and overall product activity; every
reported 90\% confidence interval excludes zero. Basic improves over
Production Two Tower, while the Advanced launch package provides
additional gains over Basic.

\begin{table}[t]
  \caption{Online lifts (\%): point estimate
  $\pm$ 90\% confidence-interval half-width.}
  \label{tab:online}
  \centering
  \footnotesize
  \setlength{\tabcolsep}{3pt}
  \begin{tabularx}{\columnwidth}{@{}Xccc@{}}
    \toprule
    Setup & E-task & C-task & Topline \\
    \midrule
    Basic vs. Two Tower
      & $+0.734\!\pm\!0.109$
      & $+0.288\!\pm\!0.035$
      & $+0.108\!\pm\!0.019$ \\
    Advanced launch package vs. Basic
      & $+0.941\!\pm\!0.154$
      & $+0.562\!\pm\!0.038$
      & $+0.174\!\pm\!0.022$ \\
    \bottomrule
  \end{tabularx}
\end{table}

\section{Related Work and Limitations}
\par\indent\textbf{Related work.}
Prior pre-rankers use reusable two-tower representations~\cite{huang2013dssm,li2022inttower},
richer tower interactions~\cite{xiong2025learnable,yang2025hit,su2023beyond},
or compressed target attention~\cite{li2025tarq}; long-sequence and generative
rankers model candidate-conditioned histories~\cite{pi2020sim,chang2023twin,si2024twinv2,han2025mtgr}.
TESR builds on HSTU sequence modeling~\cite{zhai2024hstu}, applies
request-only optimization principles~\cite{guo2025roo}, and uses
SilverTorch GPU serving~\cite{xue2025silvertorch} for EIM. It
combines these foundations with candidate-conditioned attention
while preserving reusable computation at ESR scale.
\par\indent\textbf{Limitations.}
Our evaluation uses proprietary production data and infrastructure,
which limits direct reproduction. Although the offline ablations
isolate the modeling components, the service replay evaluates the
end-to-end Basic configuration rather than individual serving
optimizations. 
\section{Conclusion}
Candidate-dependent sequence modeling in ESR is a joint modeling
and systems problem because per-candidate interactions multiply
across broad candidate pools. TESR combines explicit exact matching
and latent candidate-conditioned interaction with request-shared
history encoding, retained two-tower computation, and optimized
serving. Controlled offline ablations show predictive gains from
both interaction paths, while production-derived replay shows that
Basic sustains higher full-service throughput than Production Two
Tower under the evaluated constraints. Under live production
traffic, controlled A/B experiments show gains for Basic over
Production Two Tower and for a launch package centered on TESR
Advanced over Basic. Together, the quality, serving, and deployment
evidence establishes model--serving co-design as a practical route
to interaction-rich early-stage ranking.
\clearpage
\section*{Ethical Considerations}
\par\indent\textbf{Data and privacy.}
TESR personalizes ranking using on-platform interaction histories, including
engagement actions and the author identifiers of engaged items. These signals
can reveal behavioral preferences and therefore require appropriate controls
over access, retention, and use. Offline experiments use governed production
logs, and no user-level data, model weights, or production logs are released
with this paper. TESR does not require demographic attributes as model inputs.
\par\indent\textbf{Feedback loops and content exposure.}
Conditioning ranking more strongly on past behavior can reinforce historical
exposure patterns. EIM may particularly favor candidates from authors with
whom a user has repeatedly engaged, potentially narrowing content exposure and
making it harder for new or less-established creators to reach later ranking
stages. This concern is relevant in ESR because its outputs determine which
candidates remain available to downstream rankers.
TESR caps each action-specific exact-match count at $M$, limiting the
contribution of repeated matches within EIM. This cap does not eliminate
exposure-concentration risk, including effects arising through the learned
sequence interaction. We do not report content-diversity or creator-exposure
metrics, so the distributional effects of TESR remain a limitation. Deployment
should therefore retain downstream diversity and integrity controls and monitor
creator exposure, content diversity, and negative-feedback signals.
\par\indent\textbf{Compute cost.}
Under the same latency constraint, TESR Basic sustains 3.1\% more requests
than Production Two Tower, indicating improved serving efficiency for the
measured workload. Training throughput nevertheless falls by 17--25\%, and
model-only inference throughput also decreases. We therefore do not claim a net
reduction in energy or environmental impact, which depends on deployment scale,
hardware utilization, and retraining frequency.
\bibliographystyle{ACM-Reference-Format}
\bibliography{references}
\end{document}